\documentclass{article}

\usepackage[preprint]{neurips_2026}
\setcitestyle{numbers,square,comma,sort&compress}
\usepackage[utf8]{inputenc}
\usepackage[T1]{fontenc}
\usepackage{hyperref}
\usepackage{url}
\usepackage{booktabs}
\usepackage{amsfonts}
\usepackage{amsmath}
\usepackage{nicefrac}
\usepackage{microtype}
\usepackage{xcolor}
\usepackage{graphicx}
\usepackage{multirow}

\newcommand{\ie}{\textit{i}.\textit{e}.}

\def\algorithmname{SEGA3D}

\title{Segment and Select: Vision-Language Segmentation in 3D Scenarios}

\author{
  Yulin Chen\textsuperscript{1,2}, Zhihang Zhong\textsuperscript{3}, Yuenan Hou\textsuperscript{1 *}
  \vspace{0.3em} \\
  \textsuperscript{1}Shanghai AI Laboratory \textsuperscript{2}University of Science and Technology of China \\
  \textsuperscript{3} Shanghai Jiaotong University \\
  {\{chenyulin, houyuenan\}@pjlab.org.cn}
}

\begin{document}

\maketitle

{\renewcommand{\thefootnote}{\fnsymbol{footnote}}\footnotetext[1]{The corresponding author. This work is performed during the internship of Yulin Chen at Shanghai AI Laboratory.}}

\begin{abstract}

3D vision-language segmentation aims to segment target objects in 3D scenarios according to the linguistic instructions and visual observations. Prior art heavily relies on the coarse superpoint representation to reduce the computation complexity, which suffers from poor segmentation quality and messy object boundaries. In this paper, we propose the SEGment-And-select (\algorithmname) paradigm for 3D vision-language segmentation that directly operates on the fine-grained visual information and is free from the superpoint dependency. Specifically, we first leverage a mask candidate generator to provide fine-grained categorical mask candidates, substantially improving the quality of candidate masks over the superpoint counterparts. Then, a Large Language Model (LLM) is utilized to generate the semantic and spatial information based on the linguistic description and visual features. The LLM output and visual features are fed to the Semantic-Spatial Selector (SSS) to produce the top-ranking mask candidates. Eventually, the Loopback Verification Module (LVM) is designed to yield the segmentation mask from the selected candidate masks. Our \algorithmname~attains competitive performance on ScanRefer, ScanNet and Matterport3D benchmarks. Notably, our \algorithmname~surpasses the top-performing counterpart by \textbf{8.3} mIoU and \textbf{5.3} mIoU on ScanNet and Matterport3D, respectively. Codes will be available upon publication.

\end{abstract}

\vspace{-10pt}
\section{Introduction}
\label{sec:introduction}
\vspace{-5pt}

The objective of 3D vision-language segmentation is to segment target objects in 3D scenes according to natural-language instructions and visual observations.
As opposed to conventional 3D segmentation, this task requires a model to jointly perform language understanding, spatial grounding, and point-level mask prediction.
It covers both 3D referring segmentation, where the target is explicitly described by category, appearance, or spatial relation, and 3D reasoning segmentation, where the target must be inferred from implicit cues such as object function, commonsense knowledge, or scene intent~\cite{Scanrefer,TGNN,X-refseg3d,3d-stmn,Reason3d}.
Accurate vision-language segmentation of the 3D scenarios plays a pivotal role in embodied AI, robotics, and interactive 3D scene understanding, where agents need to interpret human instructions and ground them to precise 3D regions~\cite{3dllm,ll3da,Scanqa,Pointllm}.

\begin{figure*}[t]
    \centering
    \includegraphics[width=0.8\textwidth]{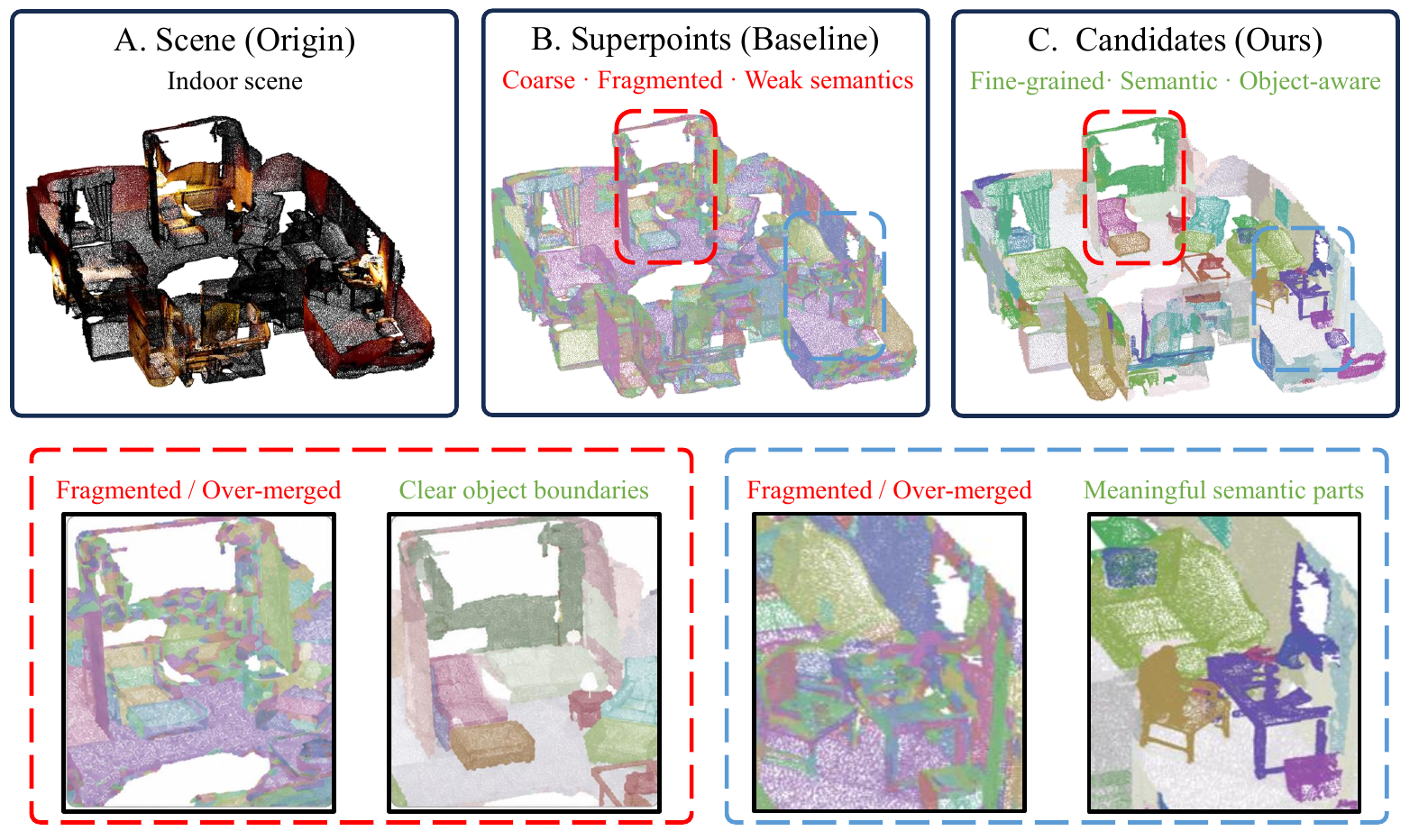}
    \vspace{-3pt}
    \caption{
    Visual comparison between the superpoint representation and our candidate representation. The superpoint representation suffers from poor semantic quality and messy object boundaries. Instead, our candidate representation provides clearer object boundaries and more accurate semantic regions. Different colors denote distinct categories.
    }
    \label{fig:motivation}
    \vspace{-10pt}
\end{figure*}

Despite recent strides in this field, 3D vision-language segmentation is still confronted with a fundamental tension between efficient target search and fine-grained mask prediction.
Contemporary methods heavily rely on superpoints or coarse region partitions to reduce the computational cost of processing large-scale point clouds~\cite{Spg,Spu-net,3d-stmn,Reason3d}.
Although such representation provide compact units for 3D-language alignment, it suffers from poor semantic quality and messy object boundaries.
As shown in Fig.~\ref{fig:motivation}, the superpoint representation fragments coherent object regions, over-merges unrelated semantic parts, and produces ambiguous object boundaries.
In contrast, our candidate representation provides clearer object boundaries and more meaningful category-aware regions.
This observation motivates us to construct an explicit candidate bank as a structured hypothesis space for language-conditioned 3D segmentation.

Recent LLM-based reasoning segmentation methods introduce task-specific language tokens, such as semantic tokens and spatial tokens, to connect language reasoning with mask decoding~\cite{Lisa,Reason3d,Pixellm}.
This token-guided paradigm enables models to infer targets from complex instructions and predict segmentation masks.
However, existing formulations typically couple target search, mask generation, and final decision into a single decoding process.
As a consequence, target localization and mask quality are largely determined by one-pass decoding from dense point-level or superpoint-level features, while explicit object-level hypothesis comparison remains under-explored.
Moreover, once a mask is decoded, its quality is seldom verified against alternative candidate masks.
These limitations suggest that 3D vision-language segmentation would benefit from a more structured formulation: the model should first construct candidate hypotheses from the 3D scene, then ground the language-inferred target in the candidate space, and finally verify the refined masks before making the final prediction.

To this end, we propose the SEGment-And-select paradigm (\algorithmname), a candidate-guided framework for 3D vision-language segmentation.
Instead of directly decoding the final mask from dense point-level or superpoint-level features, we formulate the task as candidate-level grounding followed by mask-level verification.
Given an input point cloud, our framework first constructs a fine-grained candidate bank with an upstream Candidate Generator and extracts point-wise features with a Point Encoder.
Candidate Fusion then integrates point-wise features and candidate geometry into candidate-level representations.
These representations interact with the LLM through learnable queries, allowing the model to obtain a \texttt{[SEG]}-based semantic representation and a \texttt{[LOC]}-based spatial representation.
Based on these cues, the Semantic-Spatial Selector selects the Top-\(K\) candidate hypotheses by jointly modeling semantic relevance and spatial consistency.
Finally, the Loopback Verification module refines the selected candidate masks within local regions and reranks the refined masks to yield the final mask prediction.

This candidate-level search-and-verification design provides a more explicit interface between language reasoning and 3D mask prediction.
The candidate bank supplies an object-level hypothesis space that reduces the dependence on coarse superpoint partitions.
The Semantic-Spatial Selector performs target grounding in this hypothesis space by combining semantic and spatial cues from the LLM.
Loopback Verification further decouples initial candidate grounding from the final mask decision by refining candidate masks and selecting the final output based on mask-aware verification scores.
Therefore, the final prediction is not solely determined by proposal-level scores, but also by the refined point-level mask evidence.

We evaluate \algorithmname~on ScanRefer, ScanNet and Matterport3D~\cite{scannet,matterport3d} benchmarks, and our method achieves competitive performance across all benchmarks.
Compared with Reason3D~\cite{Reason3d}, \algorithmname~improves Acc@0.25, Acc@0.50, and mIoU by \textbf{5.51}, \textbf{6.36}, and \textbf{8.38} points on ScanNet, and by \textbf{2.11}, \textbf{10.19}, and \textbf{5.34} points on Matterport3D, respectively.

Our main contributions are summarized as follows:
\begin{itemize}
    \item We reformulate 3D vision-language segmentation as the segment-and-select paradigm, which performs candidate-level grounding and mask-level verification sequentially.
    \item We construct a fine-grained candidate bank with Candidate Fusion, representing object-level mask hypotheses using point-wise features and candidate geometry.
    \item We introduce a Semantic-Spatial Selector that combines \texttt{[SEG]}-based semantic cues and \texttt{[LOC]}-based spatial cues for candidate-level grounding.
    \item We design Loopback Verification with Candidate Mask Refinement and Candidate Mask Reranking. Competitive performance on ScanRefer, ScanNet, and Matterport3D benchmarks underscores the effectiveness of our \algorithmname~.
\end{itemize}

\vspace{-8pt}
\section{Related Work}
\vspace{-5pt}
\paragraph{Point Cloud Segmentation.}
Point cloud segmentation has been widely explored and recent studies can be categorized based on scene representations~\cite{pointnet,pointnet++,Pointtransformer,ptv3,hexplane,nerf-det++,taseg,pvkd}.
Point-based representatives, such as PointNet~\cite{pointnet} and PointNet++~\cite{pointnet++}, directly learn from unordered point sets. Point Transformer~\cite{Pointtransformer} and PTv3~\cite{ptv3} improve contextual modeling and scalability through the transformer architectures.
Other works reorganize unordered point clouds into structured representations to relieve the computational complexity, including superpoint-based methods such as SPG~\cite{Spg} and SPFormer~\cite{spformer}, voxel-based methods such as MinkowskiNet~\cite{Minkowski}, and query-based mask prediction frameworks such as Mask3D~\cite{Mask3d} and OneFormer3D~\cite{kolodiazhnyi2024oneformer3d}.
These developments show that modern 3D segmentation models can provide strong point-wise features and instance- or region-level mask hypotheses, which serve as the basis for providing categorical mask candidates in our framework.

\paragraph{3D Visual Grounding and Referring Segmentation.}
3D visual grounding and referring segmentation aim to localize or segment objects in 3D scenes according to open-vocabulary labels or natural-language descriptions ~\cite{Openscene,Openmask3d,Scanrefer,moe3d,3dvg-lf}.
Early benchmarks such as ScanRefer~\cite{Scanrefer} and ReferIt3D~\cite{Referit3d} study language-conditioned object localization and instance disambiguation in real 3D scenes.

Subsequent methods improve target grounding by leveraging object proposals, candidate relations, or text-region correspondences, including InstanceRefer~\cite{Instancerefer}, 3DVG-Transformer~\cite{3dvg}, 3D-SPS~\cite{3d-sps}, TGNN~\cite{TGNN}, X-RefSeg3D~\cite{X-refseg3d}, and 3D-STMN~\cite{3d-stmn}.
These works show the effectiveness of proposal- or region-level reasoning for explicit referring expressions, but they mainly handle targets directly specified by language and do not explicitly verify refined mask quality for implicit reasoning queries.

\paragraph{Reasoning Segmentation with Large Language Models.}
Reasoning segmentation requires models to infer the target from implicit instructions involving object function, commonsense knowledge, or scene intent, and then produce the corresponding segmentation mask.
In 2D vision tasks, LISA~\cite{Lisa} introduces a \texttt{[SEG]} token to connect LLM reasoning with mask decoding, while PixelLM~\cite{Pixellm} and VISA~\cite{Visa} extend this paradigm to multi-object and video scenarios.
In 3D, Reason3D ~\cite{Reason3d} further introduces a \texttt{[LOC]} token and a hierarchical mask decoder to perform coarse-to-fine 3D mask prediction.

Although existing LLM-based reasoning segmentation methods effectively connect task-specific language tokens with mask decoders, they usually couple target search, mask generation, and final prediction within a single decoding process.
For 3D reasoning segmentation, we argue that this formulation leaves the structured 3D hypothesis space insufficiently exploited: the model should first construct target hypotheses from the scene, then ground the inferred target among these hypotheses, and finally verify the predicted mask based on its refined content.
Inspired by the perception-cognition decoupling principle~\cite{Deris}, we reformulate 3D reasoning segmentation as a candidate-level search-and-verification process.
\vspace{-8pt}
\section{Methodology}
\label{sec:methodology}
\vspace{-5pt}

\begin{figure*}[t]
    \centering
    \includegraphics[width=\textwidth]{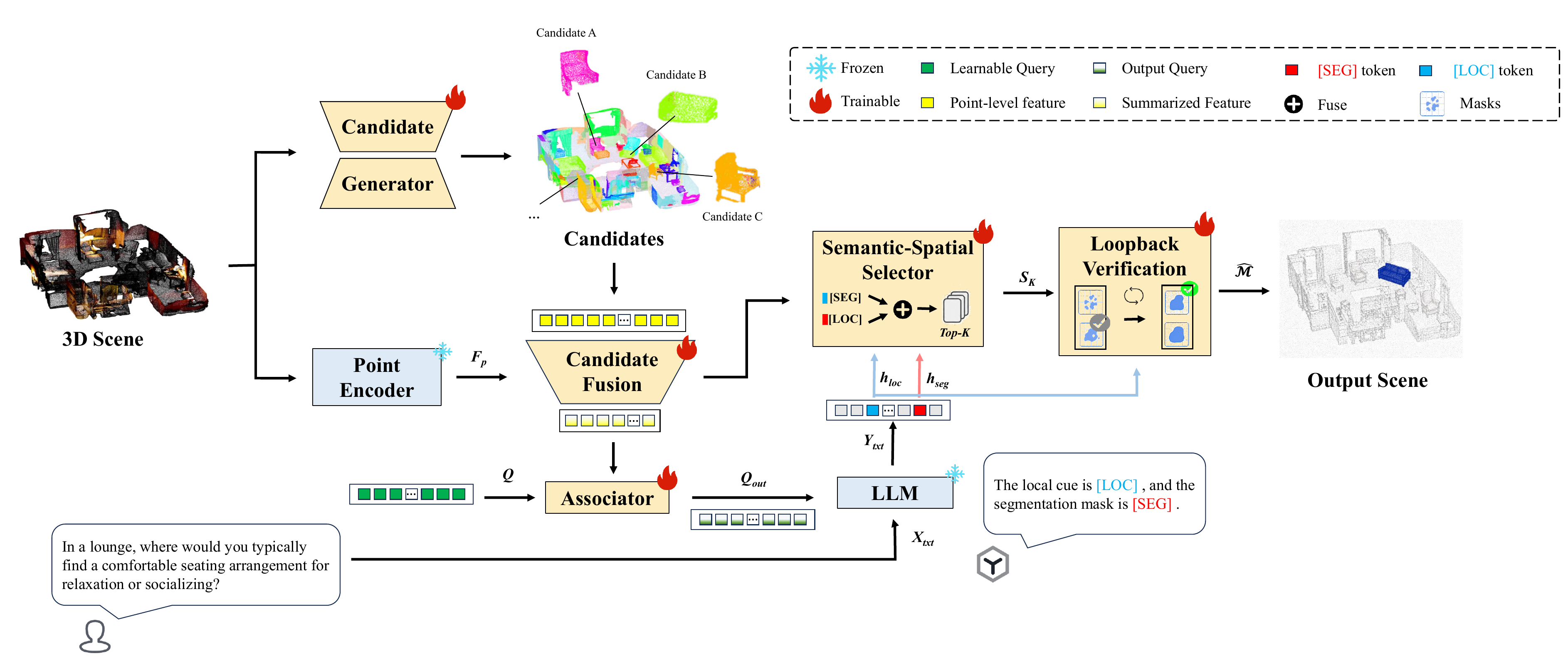}
    \vspace{-7pt}
    \caption{
    Framework overview. The model constructs a candidate bank from the input point cloud, fuses point-wise features and candidate geometry into candidate-level features, interacts with the LLM through learnable queries, selects Top-\(K\) candidates using semantic and spatial cues from \texttt{[SEG]} and \texttt{[LOC]}, and finally refines and verifies candidate masks through Loopback Verification.
    }
    \label{fig:overview}
    \vspace{-10pt}
\end{figure*}

Figure~\ref{fig:overview} provides a framework overview of our \algorithmname~.
Given an input point cloud and a textual instruction, we first leverage a candidate generator and a point encoder to produce object-level mask candidates and point-wise features, respectively. Then, the candidate fusion module fuses point-wise features and candidate geometric information, thus generating candidate-level representations. The candidate-level representations interact with the LLM through learnable queries.
The LLM then produces \(\texttt{[SEG]}\)-based semantic representation and \(\texttt{[LOC]}\)-based spatial representation, which are used for semantic-spatial candidate selection.
Finally, the selected Top-\(K\) candidates are refined and verified by the Loopback Verification module, yielding the final predicted object mask. In the following sections, we will explain each module in detail. 

\vspace{-4pt}
\subsection{Candidate Bank Construction and Representation}
\label{sec:candidate_construction}
\vspace{-2pt}
The first step is to construct a structured hypothesis space for target search.
Given an input point cloud \(\mathcal{P}\in\mathbb{R}^{N_p\times 6}\), where each point contains its 3D coordinate \((x,y,z)\) and color information \((r,g,b)\), we first use a Candidate Generator to generate a set of object-level mask candidates from the scene.
This Candidate Generator is implemented by an independently trained 3D segmentation model, which provides spatially meaningful candidate masks for the subsequent reasoning module~\cite{ptv3}.

Specifically, the Candidate Generator takes the input point cloud \(\mathcal{P}\) and generates \(N_c\) candidate regions:
\begin{equation}
    \mathcal{C}
    =
    \Phi_{\mathrm{gen}}(\mathcal{P})
    =
    \{c_i\}_{i=1}^{N_c},
    \qquad
    c_i\subseteq\{1,\dots,N_p\}.
\end{equation}
Here, \(\Phi_{\mathrm{gen}}\) denotes the pretrained Candidate Generator, and each \(c_i\) represents the point-index set of the \(i\)-th candidate region.

To obtain candidate-level representations for language-conditioned selection, we use a sparse-convolutional Point Encoder~\cite{Spu-net} \(E_{\mathrm{pc}}\) to extract point-wise features  \(F_p=E_{\mathrm{pc}}(\mathcal{P})=\{f_{p,n}\}_{n=1}^{N_p}\), and aggregate them according to the candidate masks.

For each candidate \(c_i\), we collect the corresponding point features according to its point-index set and fuse them together with candidate geometry. The candidate geometric descriptor \(g_i\) is computed from the points inside \(c_i\), including its center, spatial extent, bounding box, and position-encoded box representation. The Candidate Fusion then aggregates the internal point features and geometry of each candidate into a compact candidate-level representation:
\begin{equation}
    \{z_i\}_{i=1}^{N_c}
    =
    \Phi_{\mathrm{fus}}
    \left(
    F_{p},
    \mathcal{P},
    \mathcal{C}
    \right),
\end{equation}
where \(z_i\) denotes the fused representation of the \(i\)-th candidate.
The candidate bank and features are then used for 3D-language interaction and candidate selection.
\vspace{-4pt}
\subsection{3D-Language Interaction}
\label{sec:language_interaction}
\vspace{-2pt}
After obtaining the candidate-level representations \(\{z_i\}_{i=1}^{N_c}\), the model needs to convert the structured 3D candidate information into visual context that can be consumed by the language model.
Following the interaction design of Q-Former, we introduce a set of learnable queries \(Q\) as 3D scene queries~\cite{Blip-2}.
The Associator takes \(Q\) and the candidate-level features \(\{z_i\}_{i=1}^{N_c}\) as input, allowing the queries to aggregate scene-aware information from the candidate space and produce compact output queries \(Q_{\mathrm{out}}\).

Compared with directly feeding all point-level features into the language model, this interaction reduces the number of 3D tokens while preserving object-level structural information.
The output queries \(Q_{\mathrm{out}}\) are then fed into the language model together with the text instruction \(X_{\mathrm{txt}}\), producing the response sequence:
\begin{equation}
    Y_{out}
    =
    \mathrm{LLM}
    ([Q_{\mathrm{out}},X_{\mathrm{txt}}]).
\end{equation}
In the response sequence, the model is supervised to generate two task-related tokens: \(\texttt{[SEG]}\) and \(\texttt{[LOC]}\)~\cite{Lisa,Reason3d}.
We extract the corresponding hidden representations of these two tokens from the last hidden states of the language model, and denote them as \(h_{\mathrm{seg}}\) and \(h_{\mathrm{loc}}\), respectively.
Here, \(h_{\mathrm{seg}}\) provides semantic cues about the target category, function, or implicit intent, while \(h_{\mathrm{loc}}\) provides spatial localization cues for candidate disambiguation and local mask refinement.

\vspace{-4pt}

\vspace{-4pt}
\subsection{Semantic-Spatial Candidate Selection}
\label{sec:candidate_selection}
\vspace{-2pt}

\begin{figure*}[t]
    \centering
    \includegraphics[width=\textwidth]{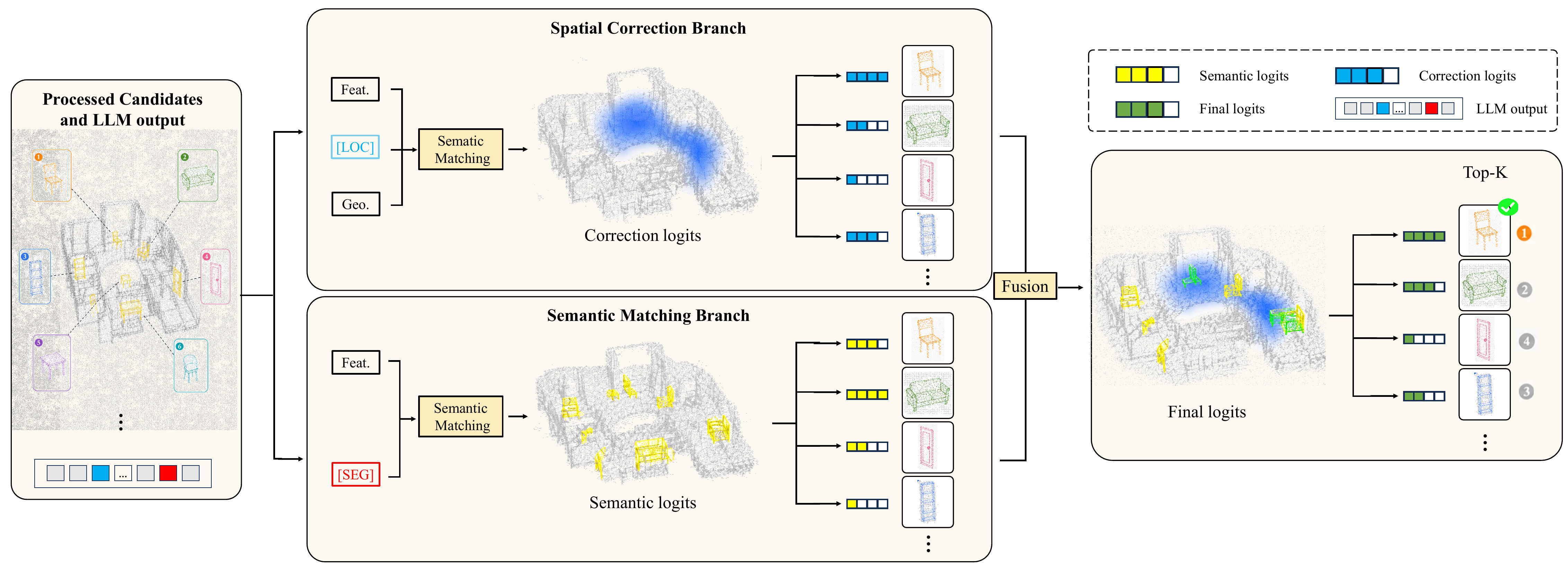}
    \caption{
    Illustration of the Semantic-Spatial Selector.
    The semantic matching branch produces base logits using candidate features and the \texttt{[SEG]} representation, while the spatial correction branch uses the \texttt{[LOC]} representation and candidate geometry to generate correction logits.
    The fused logits are used to select the Top-\(K\) candidates for constructing \(S_K\).
    }
    \label{fig:selector}
\end{figure*}

After constructing the candidate bank, the key step is to ground the target described by the language instruction in the candidate space.
In complex indoor scenes, multiple candidates may be semantically plausible, especially when they belong to the same category, have similar functions, or appear in spatially adjacent regions.
To better exploit both semantic and spatial evidence from the LLM, we propose a Semantic-Spatial Selector, which jointly models semantic relevance and spatial consistency at the candidate level.

Given candidate-level features \(\{z_i\}_{i=1}^{N_c}\), candidate geometry \(\{g_i\}_{i=1}^{N_c}\), and the language-derived semantic and spatial representations \(h_{\mathrm{seg}}\) and \(h_{\mathrm{loc}}\), the selector performs joint candidate filtering:
\begin{equation}
    \mathcal{C}_{K}
    =
    \Phi_{\mathrm{sel}}
    \left(
    \{(z_i,g_i)\}_{i=1}^{N_c},
    h_{\mathrm{seg}},
    h_{\mathrm{loc}}
    \right).
\end{equation}
Here, \(\mathcal{C}_{K}\) denotes the retained Top-\(K\) candidate regions.

For subsequent Loopback Verification, we further construct a selected candidate-feature set:
\begin{equation}
    S_{K}
    =
    \left\{
    (c_i,z_i,g_i,F_{p,i})
    \mid
    c_i\in\mathcal{C}_{K}
    \right\},
\end{equation}
where \(F_{p,i}=\{f_{p,n}\mid n\in\Omega(c_i)\}\) denotes the local point-level features associated with candidate \(c_i\), and \(\Omega(c_i)\subseteq\{1,\dots,N_p\}\) denotes the point-index set of the local region cropped around the candidate.
Here, \(\mathcal{C}_{K}\) contains the selected candidate regions, while \(S_{K}\) augments them with candidate-level features, geometric descriptors, and cropped local point-level features.

Figure~\ref{fig:selector} illustrates the design of the Semantic-Spatial Selector.
The semantic matching branch uses \(h_{\mathrm{seg}}\) to measure the semantic consistency between each candidate and the target described by the instruction, producing base logits for candidate recall.
Built on these semantic scores, the spatial correction branch incorporates \(h_{\mathrm{loc}}\) and candidate geometry \(g_i\) to calibrate candidates with spatial evidence.
The fused logits are then used to select the Top-\(K\) candidates, and the corresponding \(S_{K}\) is passed to Loopback Verification.

\vspace{-4pt}
\subsection{Loopback Verification}
\label{sec:loopback_verification}
\vspace{-2pt}

The Semantic-Spatial Selector provides the selected candidate-feature set \(S_{K}\).
Directly using the top-ranked candidate would make the final prediction mainly depend on proposal-level scores.
Therefore, we introduce Loopback Verification to refine the Top-\(K\) candidate masks within local regions and rerank the refined results.

\begin{figure*}[t]
    \centering
    \includegraphics[width=\textwidth]{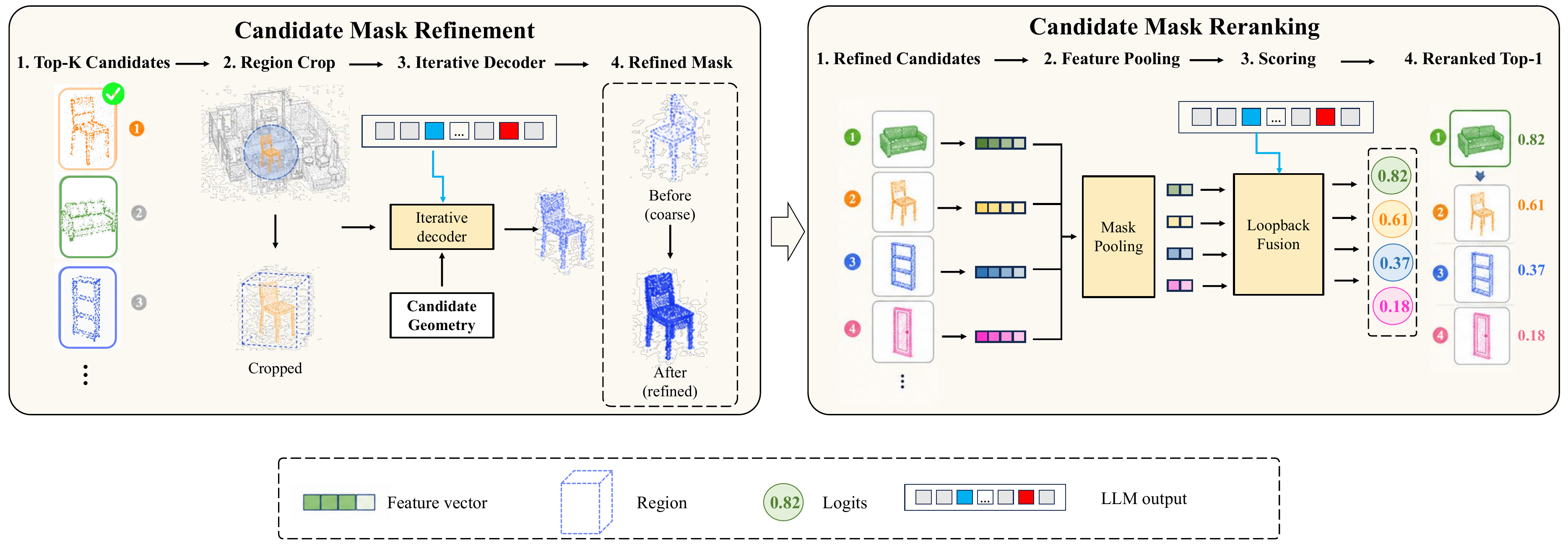}
    \vspace{-2pt}
    \caption{
    Illustration of Loopback Verification.
    Candidate Mask Refinement refines each selected Top-\(K\) candidate into a point-level mask using local point-level features and the \texttt{[LOC]}-derived localization cue.
    Candidate Mask Reranking then re-pools mask-aware features and predicts mask-level verification scores to select the final output.
    }
    \label{fig:reloop}
    \vspace{-3pt}
\end{figure*}

As illustrated in Fig.~\ref{fig:reloop}, Loopback Verification contains two stages.
Candidate Mask Refinement uses \(F_{p,i}\), candidate geometry, and the localization prior derived from \(h_{\mathrm{loc}}\) to refine the point-level mask within the cropped local region.
Candidate Mask Reranking then re-pools features according to the refined mask to obtain a mask-aware feature \(\tilde{z}_i\), and predicts a mask-level verification score using \(\tilde{z}_i\), candidate geometry, and \(h_{\mathrm{loc}}\).
The overall process is summarized as:
\begin{equation}
    \{(\hat{m}_i,s_i^{\mathrm{loop}})\}_{i=1}^{K}
    =
    \Phi_{\mathrm{loop}}
    \left(
    S_{K},
    h_{\mathrm{loc}}
    \right),
\end{equation}
where \(\hat{m}_i\) denotes the refined point-level mask of the \(i\)-th selected candidate, and \(s_i^{\mathrm{loop}}\) denotes its mask-level verification score.

Finally, we select the refined mask with the highest verification score:
\begin{equation}
    \hat{\mathcal{M}}
    =
    \hat{m}_{i^{*}},
    \qquad
    i^{*}
    =
    \arg\max_{i\in\{1,\dots,K\}}
    s_i^{\mathrm{loop}}.
\end{equation}
In this way, Loopback Verification makes the final prediction based on both candidate-level relevance and refined mask evidence.
\vspace{-8pt}
\subsection{Training Objective}
\vspace{-2pt}
The training objective consists of three terms, \ie, language generation, candidate selection, and loopback verification:
\[
\mathcal{L}
=
\mathcal{L}_{\mathrm{out}}
+
\lambda_{\mathrm{sel}}\mathcal{L}_{\mathrm{sel}}
+
\lambda_{\mathrm{loop}}
\left(
\mathcal{L}_{\mathrm{mask}}
+
\mathcal{L}_{\mathrm{score}}
\right).
\]
The language loss \(\mathcal{L}_{\mathrm{out}}\) follows the standard auto-regressive cross-entropy objective and supervises the language-output pathway to generate the response sequence with valid \texttt{[SEG]} and \texttt{[LOC]} tokens. The candidate selection loss \(\mathcal{L}_{\mathrm{sel}}\) supervises the Semantic-Spatial Selector to identify the target candidate from the candidate bank. Candidate labels are assigned according to the IoU between each candidate and the ground-truth mask. For loopback verification, \(\mathcal{L}_{\mathrm{mask}}\) denotes the mask-related objective, including refined point-level mask supervision and local region prior supervision with BCE, Dice, and Lovasz losses. \(\mathcal{L}_{\mathrm{score}}\) supervises the mask-level verification score so that the refined mask corresponding to the target candidate receives the highest score. Detailed definitions and implementation settings are provided in the appendix.
\vspace{-10pt}
\section{Experiments}
\label{sec:experiments}
\vspace{-5pt}

\subsection{Experimental Setup}
\label{sec:exp_setup}

\paragraph{Tasks \& datasets.}
We evaluate our \algorithmname~on two tasks, \ie, 3D referring segmentation and 3D reasoning segmentation.
For 3D referring segmentation, we train models on ScanRefer~\cite{Scanrefer} and report results on its validation set.
For 3D reasoning segmentation, we follow the dataset and evaluation protocol of Reason3D~\cite{Reason3d}, and evaluate on ScanNet~\cite{scannet} and Matterport3D~\cite{matterport3d}.
These two tasks evaluate explicit language grounding and implicit reasoning-based 3D target segmentation, respectively.
\vspace{-5pt}
\paragraph{Evaluation metrics.}
Following prior work~\cite{Reason3d}, we report Acc@0.25, Acc@0.50, and mean Intersection-over-Union (mIoU).
Acc@$k$ measures the percentage of queries whose predicted mask has an IoU larger than \(k\) with the ground-truth mask, while mIoU measures the average mask overlap over all samples.
\vspace{-5pt}
\paragraph{Candidate generation \& inference.}
We use pretrained PTv3~\cite{ptv3} segmentation models built on Pointcept as upstream Candidate Generators.
They produce object-level proposals, which are filtered and merged into the final candidate bank.
All tasks follow the same candidate-guided inference pipeline: the model constructs a candidate bank, performs Candidate Fusion, filters candidates with the Semantic-Spatial Selector, and applies Loopback Verification to the selected Top-\(K\) candidates.
\vspace{-5pt}
\paragraph{Implementation details.}
We use Flan-T5-XL as the language model and a pretrained sparse U-Net-style backbone as the Point Encoder.
The Candidate Generator, Point Encoder, and LLM are kept frozen, while Candidate Fusion, the Associator, the Semantic-Spatial Selector, and Loopback Verification are optimized.
We train the model for 100 epochs using the AdamW optimizer, mixed-precision training, and a batch size of 24 per GPU on 8 NVIDIA H200 GPUs.
During inference, we set \(K=4\) for Loopback Verification and select the refined mask with the highest verification score as the final prediction.
During training, we mix a small portion of ground-truth candidates to stabilize candidate selection and Loopback Verification, while no ground-truth candidates are used during inference.
More implementation details are provided in Appendix~\ref{app:implementation_details}.

\vspace{-7pt}

\subsection{Quantitative Comparison}

\textbf{3D Reasoning Segmentation.}
We first evaluate the proposed method on 3D reasoning segmentation.
This task requires the model to infer the target from implicit cues such as object function, commonsense knowledge, or scene context, and then produce the corresponding point-level mask.

\begin{table*}[t]
\centering
\caption{3D reasoning segmentation results on ScanNet and Matterport3D.}
\label{tab:reason_main}
\resizebox{\textwidth}{!}{
\begin{tabular}{llcccccc}
\toprule
\multirow{2}{*}{Method}
& \multirow{2}{*}{Venue}
& \multicolumn{3}{c}{ScanNet}
& \multicolumn{3}{c}{Matterport3D} \\
\cmidrule(lr){3-5}
\cmidrule(lr){6-8}
& & Acc@0.25 & Acc@0.50 & mIoU
& Acc@0.25 & Acc@0.50 & mIoU \\
\midrule
OpenScene~\cite{Openscene} & CVPR'23 & 4.22 & 0.97 & 5.03 & 4.07 & 0.57 & 6.36 \\
OpenScene~\cite{Openscene} + Flan-T5~\cite{t5} & CVPR'23 & 24.68 & 7.14 & 15.03 & 19.98 & 4.02 & 13.60 \\
OpenMask3D~\cite{Openmask3d} & NeurIPS'23 & 5.70 & 3.25 & 7.14 & 3.25 & 0.12 & 5.96 \\
OpenMask3D~\cite{Openmask3d} + Flan-T5~\cite{t5} & NeurIPS'23 & 20.78 & 6.82 & 13.38 & 17.46 & 0.23 & 9.07 \\
3D-STMN~\cite{3d-stmn} & AAAI'24 & 25.43 & 17.78 & 18.23 & 20.68 & 10.81 & 13.47 \\
Llama2~\cite{llama2} + CLIP~\cite{clip} & Baseline & 39.26 & 25.93 & 27.23 & 28.51 & 14.86 & 17.80 \\
Reason3D~\cite{Reason3d} & 3DV'25 & 43.21 & 32.10 & 31.20 & 31.22 & 17.43 & 19.54 \\
\midrule
Ours & -- & \textbf{48.72} & \textbf{38.46} & \textbf{39.58} & \textbf{33.33} & \textbf{27.62} & \textbf{24.88} \\
\bottomrule
\end{tabular}
}
\vspace{-5pt}
\end{table*}

Table~\ref{tab:reason_main} reports the results on ScanNet and Matterport3D.
Our method achieves the best performance on both datasets across all evaluation metrics.
Compared with Reason3D, our method improves Acc@0.25, Acc@0.50, and mIoU by \textbf{5.51}, \textbf{6.36}, and \textbf{8.38} points on ScanNet, respectively. On Matterport3D, our method improves Acc@0.25, Acc@0.50, and mIoU by \textbf{2.11}, \textbf{10.19}, and \textbf{5.34} points, respectively.
Compared with open-vocabulary segmentation and referring segmentation baselines, our method better handles queries that require functional or commonsense reasoning.
Compared with Reason3D, the gains indicate that candidate-level grounding and Loopback Verification over refined masks improve the reliability of final mask prediction.
The candidate bank provides structured object-level mask hypotheses, the Semantic-Spatial Selector performs target grounding in this hypothesis space, and Loopback Verification further verifies the refined candidates for final selection.

\textbf{3D Referring Segmentation.}
\label{sec:refer_results}
We further evaluate the proposed method on the ScanRefer~\cite{Scanrefer} validation set, where the model is required to segment the target object according to an explicit natural-language description.
Table~\ref{tab:refer_main} compares our method with existing 3D grounding and referring segmentation methods.
To keep the main table concise, we report Acc@0.25, Acc@0.50, and mIoU on the Overall split, and provide detailed Unique/Multiple split results in the appendix.

As shown in Table~\ref{tab:refer_main}, our method achieves the best Overall Acc@0.50 and mIoU, while matching Reason3D on Acc@0.25.
Compared with Reason3D, our method improves Acc@0.50 and mIoU by \textbf{10.8} and \textbf{2.9} points, respectively.
These results indicate that the proposed candidate-level grounding and Loopback Verification improve mask quality while maintaining strong target localization ability.
\begin{table}[t]
\centering
\caption{3D referring expression segmentation results on the ScanRefer validation set.}
\label{tab:refer_main}
\small
\setlength{\tabcolsep}{6pt}
\begin{tabular}{llccc}
\toprule
Method & Venue & Acc@0.25 & Acc@0.50 & mIoU \\
\midrule
ScanRefer~\cite{Scanrefer} & ECCV'20 & 38.2 & 25.5 & 23.5 \\
3DVG-Transformer~\cite{3dvg} & ICCV'21 & 49.3 & 36.1 & 31.4 \\
3D-SPS~\cite{3d-sps} & CVPR'22 & 50.1 & 37.6 & 32.1 \\
3D-LLM~\cite{3dllm} & NeurIPS'23 & 31.1 & 16.3 & 17.6 \\
TGNN~\cite{TGNN} & AAAI'21 & 38.6 & 32.7 & 28.8 \\
X-RefSeg3D~\cite{X-refseg3d} & AAAI'24 & 40.3 & 33.8 & 29.9 \\
3D-STMN~\cite{3d-stmn}& AAAI'24 & 54.6 & 39.8 & 39.5 \\
Reason3D~\cite{Reason3d}& 3DV'25 & \textbf{57.9} & 41.9 & 42.0 \\
\midrule
Ours & -- & \textbf{57.9} & \textbf{52.7} & \textbf{44.9} \\
\bottomrule
\end{tabular}
\vspace{-5pt}
\end{table}

\vspace{-7pt}
\subsection{Ablation Study}
\label{sec:ablation}

We conduct ablation studies on the ScanRefer~\cite{Scanrefer} validation set and the detailed performance is summarized in Table~\ref{tab:ablation_all}.
The ablations cover \(\texttt{[LOC]}\) guidance, selector design, Candidate Fusion, candidate bank construction, Loopback Verification, and training losses.

\textbf{Effect of \(\texttt{[LOC]}\) guidance.}
Removing \(\texttt{[LOC]}\)-guided information causes a clear drop of \textbf{9.25} Acc@0.50 and \textbf{7.22} mIoU, validating the importance of spatial localization cues for candidate disambiguation and mask refinement.

\textbf{Effect of selector design.}
The full Semantic-Spatial Selector outperforms both Base-only and Corr-only variants, improving mIoU by \textbf{3.64} and \textbf{2.99} points, respectively.
The depth study further shows that a shallow selector is insufficient, while increasing the depth beyond the default setting does not bring further gains.

\textbf{Effect of Candidate Fusion.}
Replacing Candidate Fusion with mean or max pooling reduces mIoU by \textbf{2.65} and \textbf{2.05} points, respectively, showing that learnable candidate feature fusion better preserves discriminative candidate information.

\textbf{Effect of candidate bank construction.}
Using SP-UNet candidates only moderately reduces mIoU by \textbf{0.97} points, suggesting that the framework remains robust to different candidate sources.
Removing GT candidate mixing during training slightly reduces mIoU by \textbf{0.91} points, indicating that high-quality auxiliary candidates help stabilize training.

\textbf{Effect of Loopback Verification.}
Removing Loopback Verification decreases Acc@0.50 by \textbf{5.30} points.
Both Candidate Mask Refinement and Candidate Mask Reranking contribute to the full module, with larger gains coming from mask refinement.

\textbf{Effect of training losses.}
Removing local region prior supervision or using simplified loss variants consistently reduces performance.
The relatively small drops of these variants suggest that the framework is stable and not overly dependent on a single auxiliary loss.

\begin{table*}[t]
\centering
\caption{Main ablation studies on the ScanRefer~\cite{Scanrefer} validation set.}
\label{tab:ablation_all}
\scriptsize
\setlength{\tabcolsep}{5pt}
\renewcommand{\arraystretch}{0.95}

\begin{minipage}[t]{0.48\textwidth}
\centering
\begin{tabular}{lccc}
\toprule
Variant & Acc@0.25 & Acc@0.50 & mIoU \\
\midrule

\multicolumn{4}{l}{\textbf{(A) Overall LOC guidance}} \\
Ours & \textbf{57.94} & \textbf{52.73} & \textbf{44.96} \\
w/o LOC & 47.93 & 43.48 & 37.74 \\

\midrule
\multicolumn{4}{l}{\textbf{(B) Selector design(layer = 6)}} \\
Base-only selector & 52.90 & 47.43 & 41.32 \\
Corr-only selector & 53.82 & 48.86 & 41.97 \\
Selector layer = 1 & 51.81 & 47.01 & 40.56 \\
Selector layer = 3 & 55.59 & 50.21 & 43.61 \\
Selector layer = 8 & 55.84 & 50.88 & 43.64 \\

\midrule
\multicolumn{4}{l}{\textbf{(C) Candidate Fusion}} \\
Mean pooling & 54.33 & 49.20 & 42.31 \\
Max pooling & 54.84 & 49.79 & 42.91 \\

\bottomrule
\end{tabular}
\end{minipage}
\hfill
\begin{minipage}[t]{0.48\textwidth}
\centering
\begin{tabular}{lccc}
\toprule
Variant & Acc@0.25 & Acc@0.50 & mIoU \\
\midrule

\multicolumn{4}{l}{\textbf{(D) Candidate bank construction}} \\
SP-UNet candidates & 56.34 & 50.63 & 43.99 \\
w/o GT candidate mixing & 56.18 & 51.13 & 44.05 \\

\midrule
\multicolumn{4}{l}{\textbf{(E) Loopback Verification}} \\
w/o Loopback Verification & 56.69 & 47.43 & 43.87 \\
w/o Mask Refinement & 56.85 & 49.29 & 44.00 \\
w/o Mask Reranking & 57.44 & 52.06 & 44.49 \\

\midrule
\multicolumn{4}{l}{\textbf{(F) Training losses}} \\
w/o local region prior loss & 56.01 & 50.71 & 43.73 \\
w/ BCE loss only & 56.52 & 50.79 & 44.23 \\
Selector w/ CE only & 56.01 & 51.47 & 44.49\\
w/ unweighted loss & 55.86 & 50.80 & 43.83\\
\bottomrule
\end{tabular}
\end{minipage}
\end{table*}

\subsection{Visualization Results}
\label{sec:visualization}

\begin{figure*}[t]
     \centering
     \includegraphics[width=0.8\textwidth]{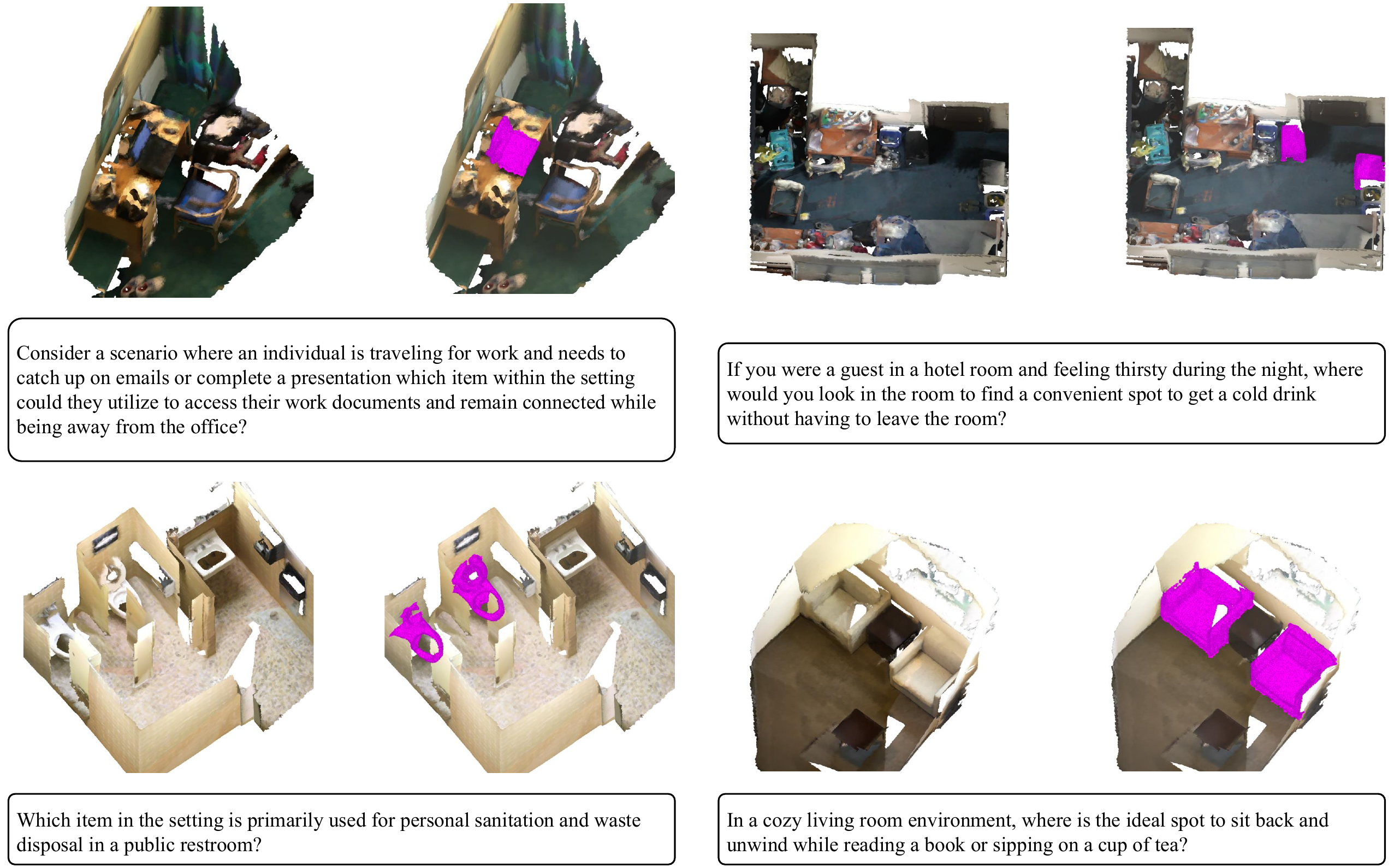}
     \vspace{-3pt}
     \caption{
     Qualitative results on 3D referring segmentation and 3D reasoning segmentation.
     Each example shows the input point cloud, text instruction, predicted mask, and ground-truth mask.
     }
    \label{fig:visualization}
    \vspace{-5pt}
\end{figure*}

Figure~\ref{fig:visualization} shows qualitative results of the proposed method on 3D referring segmentation and 3D reasoning segmentation.
Each example includes the input point cloud, the text instruction, the predicted mask, and the ground-truth mask.
The results show that the model can handle not only explicit category, appearance, or spatial relation descriptions, but also infer target objects from implicit cues such as object function, commonsense knowledge, and scene context.
In complex scenes, the Semantic-Spatial Selector helps select the correct target from multiple similar candidates, while Loopback Verification refines candidate masks and reranks the refined results for final prediction.

\section{Conclusion}
\label{sec:conclusion}
\vspace{-5pt}

In this paper, we present a candidate-guided framework for 3D vision-language segmentation, reformulating dense mask prediction as candidate-level grounding followed by mask-level verification.
The framework constructs a structured candidate bank, performs semantic-spatial candidate selection with LLM-derived cues, and refines the selected candidates through Loopback Verification.
Experiments on 3D referring and reasoning segmentation demonstrate the effectiveness of the proposed search-and-verification design, with ablation studies validating the main components.
A current limitation is that the evaluation mainly focuses on indoor scene benchmarks; extending the framework to broader scene distributions, multi-object instructions, and richer 3D vision-language tasks remains future work.

\bibliographystyle{unsrtnat}
\bibliography{reference}

\newpage
\appendix

\section{Implementation Details}
\label{app:implementation_details}

This section provides additional implementation details that are omitted from the main paper.
We mainly describe the candidate generation protocol, module designs, and training objectives.
Detailed hyperparameters and code-level configurations will be released with our implementation at GitHub upon acceptance.

Unless otherwise specified, all experiments use the same candidate-guided framework as described in the main paper.
Given an input point cloud and a textual instruction, the model first constructs a candidate bank using an upstream Candidate Generator and extracts point-wise features using a Point Encoder.
Candidate Fusion then converts each raw candidate mask into a compact candidate-level representation by aggregating internal point-wise features and geometric cues.
The candidate-level representations interact with the LLM through learnable queries in the Associator.
The semantic-rich and spatial-aware representations from \texttt{[SEG]} and \texttt{[LOC]} are then used by the Semantic-Spatial Selector to select Top-\(K\) candidate hypotheses.
Finally, Loopback Verification refines and verifies the selected candidates to produce the final object mask.

\paragraph{Compute resources.}
All experiments are conducted on 8 NVIDIA H200 GPUs with 141GB memory. The batch size is 24 per GPU. The model typically converges within 50--70 epochs, and each epoch takes approximately 9.5 minutes. Inference can be performed on a single GPU, while we use the same 8-GPU environment for convenience in evaluation.

\paragraph{Broader impacts.}
This work is a research study on 3D vision-language segmentation for indoor scene understanding.
It may benefit applications such as embodied AI, robotics, AR/VR, and assistive scene understanding by enabling more flexible language-driven object localization in 3D environments.
Since the method is evaluated only on established research datasets and is not deployed in real-world systems, we do not identify immediate societal risks.
However, future deployment of similar 3D perception systems in private indoor spaces should consider privacy, consent, and robustness to failure cases.

\paragraph{Licenses and terms of use.}
We use ScanNet, Matterport3D, ScanRefer, ScanNet200, and Reason3D under their official access procedures and terms of use. All datasets are used only for research purposes, and we cite the original papers and follow the corresponding licenses and usage policies.

\paragraph{Limitations.}
The current evaluation mainly focuses on established indoor scene benchmarks, including ScanRefer, ScanNet, and Matterport3D. Although these datasets cover diverse indoor scenes, they may not fully reflect broader 3D scene distributions or more complex multi-object language instructions. Extending the evaluation to more diverse environments and richer 3D vision-language tasks remains future work.

\paragraph{Backbone and point features.}
We decouple candidate generation from point feature extraction.
The upstream Candidate Generator is implemented with pretrained PTv3 segmentation models based on the Pointcept framework.
These models are used to generate object-level candidate masks.
For the main reasoning model, we use a separate Point Encoder to extract point-wise features from the input point cloud.
In our implementation, this Point Encoder follows a sparse U-Net-style architecture based on submanifold sparse convolutions~\cite{Spu-net}.
The extracted point-wise features are shared by Candidate Fusion, the Semantic-Spatial Selector, and Loopback Verification.

\paragraph{Training and inference.}
During training, we mix a small portion of ground-truth candidates into the candidate bank as auxiliary candidates to stabilize the learning of the Semantic-Spatial Selector and Loopback Verification modules.
This setting is used only during training and is not used during inference.
During inference, all candidates are generated by the pretrained Candidate Generator without using any ground-truth information.
For scenes with large candidate sets, the candidate-guided design allows the model to first reduce the search space through candidate selection before applying mask-level refinement and verification.

\section{Details of Candidate Generation}
\label{app:candidate_generation}

The candidate generation process is performed by pretrained Candidate Generators.
The purpose of the Candidate Generator is to convert dense search over the raw point cloud into a compact object-level candidate space.
Each candidate represents a set of points, and subsequent modules aggregate point-wise features and geometric information inside the candidate to obtain candidate-level representations.

\paragraph{Candidate generation for ScanRefer.}
For 3D referring segmentation on ScanRefer, we train PTv3 segmentation models based on the Pointcept framework under two label settings.
The first setting uses ScanNet200 categories, which cover a richer set of object classes.
The second setting uses the 20 base categories from ScanNetV2, which provide more stable segmentation for common classes.
We merge the two sets of segmentation results and filter the candidate regions according to their coverage to obtain the final candidate bank.
This design allows the candidate space to benefit from both fine-grained category coverage and stable base-category segmentation.

\paragraph{Candidate generation for reasoning segmentation.}
For 3D reasoning segmentation, we use dataset-specific Candidate Generators.
On ScanNet, candidates are generated by a PTv3 segmentation model trained on ScanNet200~\cite{scannet200}.
On Matterport3D, candidates are generated by a PTv3 segmentation model trained on Matterport3D.
Since the official Matterport3D category setting in Pointcept covers only a limited set of classes, we expand the candidate category set to improve candidate coverage.
This is important for reasoning segmentation, where the target may be described by function, commonsense, or scene context rather than by an explicit category name.

\paragraph{Ground-truth candidate mixing.}
To stabilize the learning of the Semantic-Spatial Selector and Loopback Verification modules, we mix a small portion of ground-truth candidates into the candidate bank during training.
This provides high-quality candidate hypotheses and helps the model learn candidate selection and mask refinement more effectively.
The setting is used only as training-time assistance.
During inference, no ground-truth candidates are used, and all candidates are generated by the pretrained Candidate Generators.

\section{Detailed Module Designs}
\label{app:module_details}

\subsection{Candidate Fusion}
\label{app:candidate_fusion}

In the main paper, Candidate Fusion converts raw candidate masks into candidate-level representations for language-conditioned selection.
Given the input point cloud \(\mathcal{P}\in\mathbb{R}^{N_p\times 6}\), the Candidate Generator produces a candidate bank:
\begin{equation}
    \mathcal{C}
    =
    \Phi_{\mathrm{gen}}(\mathcal{P})
    =
    \{c_i\}_{i=1}^{N_c},
    \qquad
    c_i\subseteq\{1,\dots,N_p\}.
\end{equation}
Here, \(\Phi_{\mathrm{gen}}\) denotes the pretrained Candidate Generator, and each \(c_i\) denotes the point-index set of the \(i\)-th candidate region.

We use a Point Encoder \(E_{\mathrm{pc}}\) to extract point-wise features:
\begin{equation}
    F_p
    =
    E_{\mathrm{pc}}(\mathcal{P})
    =
    \{f_{p,n}\}_{n=1}^{N_p}.
\end{equation}
For each candidate \(c_i\), we collect the corresponding point features according to its point-index set and fuse them together with candidate geometry.
The candidate geometric descriptor \(g_i\) is computed from the points inside \(c_i\), including its center, spatial extent, bounding box, and position-encoded box representation.

Candidate Fusion aggregates the internal point features and geometry of each candidate into a compact candidate-level representation:
\begin{equation}
    \{z_i\}_{i=1}^{N_c}
    =
    \Psi_{\mathrm{fus}}
    \left(
    F_p,
    \mathcal{P},
    \mathcal{C}
    \right),
\end{equation}
where \(z_i\) denotes the fused representation of the \(i\)-th candidate.

Unlike fixed mean pooling or max pooling, the learnable fusion module adaptively aggregates point-wise features according to their importance inside the candidate.
This is particularly useful when a candidate contains a limited number of points or has noisy boundaries, since fixed pooling may suppress locally discriminative information.
In this way, Candidate Fusion serves as the bridge between raw mask candidates and language-conditioned candidate reasoning.

\subsection{3D-Language Interaction}
\label{app:language_interaction}

After obtaining the candidate-level representations \(\{z_i\}_{i=1}^{N_c}\), the model needs to convert the structured 3D candidate information into visual context that can be consumed by the language model.
Following the interaction design of Q-Former, we introduce a set of learnable queries \(Q\) as 3D scene queries~\cite{Blip-2}.
The Associator takes \(Q\) and the candidate-level features \(\{z_i\}_{i=1}^{N_c}\) as input, allowing the queries to aggregate scene-aware information from the candidate space and produce compact output queries \(Q_{\mathrm{out}}\).

Compared with directly feeding all point-level features into the language model, this interaction reduces the number of 3D tokens while preserving object-level structural information.
The output queries \(Q_{\mathrm{out}}\) are then fed into the language model together with the text instruction \(X_{\mathrm{txt}}\), producing the response sequence:
\begin{equation}
    Y
    =
    \mathrm{LLM}
    ([Q_{\mathrm{out}},X_{\mathrm{txt}}]).
\end{equation}

In the response sequence, the model is supervised to produce two task-related tokens: \texttt{[SEG]} and \texttt{[LOC]}.
We extract the corresponding hidden representations of these two tokens from the last hidden states of the language model, and denote them as \(h_{\mathrm{seg}}\) and \(h_{\mathrm{loc}}\), respectively.
Here, \(h_{\mathrm{seg}}\) provides semantic cues about the target category, function, or implicit intent, while \(h_{\mathrm{loc}}\) provides spatial localization cues for candidate disambiguation and local mask refinement.

\subsection{Semantic-Spatial Selector}
\label{app:selector_details}

The Semantic-Spatial Selector grounds the target described by the instruction in the candidate space.
Given candidate-level features \(\{z_i\}_{i=1}^{N_c}\), candidate geometry \(\{g_i\}_{i=1}^{N_c}\), and the language-derived semantic and spatial representations \(h_{\mathrm{seg}}\) and \(h_{\mathrm{loc}}\), the selector performs joint candidate filtering:
\begin{equation}
    \mathcal{C}_{K}
    =
    \Phi_{\mathrm{sel}}
    \left(
    \{(z_i,g_i)\}_{i=1}^{N_c},
    h_{\mathrm{seg}},
    h_{\mathrm{loc}}
    \right).
\end{equation}
Here, \(z_i\) denotes the feature of the \(i\)-th candidate, \(g_i\) denotes its geometric information, and \(\mathcal{C}_{K}\) denotes the retained Top-\(K\) candidates.

The selector contains two complementary branches.
The semantic matching branch uses \(h_{\mathrm{seg}}\) to measure the semantic consistency between each candidate and the target described by the instruction, producing base logits for candidate recall.
Built on these semantic scores, the spatial correction branch further incorporates \(h_{\mathrm{loc}}\) and candidate geometry \(g_i\) to calibrate candidates with spatial evidence, rather than performing an independent classification.
The fused logits are then used to select the Top-\(K\) candidates.

For Loopback Verification, we construct a selected candidate-feature set:
\begin{equation}
    \mathcal{S}_{K}
    =
    \left\{
    (c_i,z_i,g_i,F_{p,i})
    \mid
    c_i\in\mathcal{C}_{K}
    \right\},
\end{equation}
where \(F_{p,i}=\{f_{p,n}\mid n\in\Omega(c_i)\}\) denotes the local point-level features associated with candidate \(c_i\), and \(\Omega(c_i)\subseteq\{1,\dots,N_p\}\) denotes the point-index set of the local region cropped around the candidate.

\subsection{Loopback Verification}
\label{app:loopback_details}

Loopback Verification is designed to address two issues.
First, the initial candidate boundaries may be inaccurate.
Second, the selector's proposal-level score may not fully reflect the actual quality of the refined mask.
Rather than directly taking the top-ranked candidate as the final prediction, Loopback Verification refines the selected Top-\(K\) candidates and reranks them based on mask-level evidence.

Specifically, for each selected candidate, we package its candidate feature, geometric descriptor, and cropped local point-level features into the selected candidate-feature set \(\mathcal{S}_{K}\).
Together with the LLM-derived localization representation \(h_{\mathrm{loc}}\), Loopback Verification outputs a refined mask and a mask-level verification score for each selected candidate:
\begin{equation}
    \{(\hat{m}_i,s_i^{\mathrm{loop}})\}_{i=1}^{K}
    =
    \Phi_{\mathrm{loop}}
    \left(
    \mathcal{S}_{K},
    h_{\mathrm{loc}}
    \right).
\end{equation}
Here, \(\hat{m}_i\) denotes the refined point-level mask of the \(i\)-th selected candidate, and \(s_i^{\mathrm{loop}}\) denotes its mask-level verification score.

During Candidate Mask Refinement, the model uses the local point-level features \(F_{p,i}\), the spatial cues from \(h_{\mathrm{loc}}\), and local region prior supervision to re-estimate the points inside and around each candidate, thereby improving object shape and boundary quality.
For Candidate Mask Reranking, we do not directly score the initial candidate feature.
Instead, the module re-pools local point-level features \(F_{p,i}\) according to the actual region covered by the refined mask and obtains a mask-aware feature:
\begin{equation}
    \tilde{z}_i
    =
    \frac{
    \sum_{n\in\Omega(c_i)}
    \hat{m}_i(n) f_{p,n}
    }{
    \sum_{n\in\Omega(c_i)}
    \hat{m}_i(n)+\epsilon
    }.
\end{equation}
This feature reflects the local content selected by the refined mask, rather than the initial candidate representation.
The verification score \(s_i^{\mathrm{loop}}\) is then predicted from the mask-aware feature, candidate geometry, and the spatial localization cue \(h_{\mathrm{loc}}\).

Finally, the refined mask with the highest verification score is selected as the final prediction:
\begin{equation}
    \hat{\mathcal{M}}
    =
    \hat{m}_{i^{*}},
    \qquad
    i^{*}
    =
    \arg\max_{i\in\{1,\dots,K\}}
    s_i^{\mathrm{loop}}.
\end{equation}
In this way, the final prediction is determined by both candidate-level relevance and refined mask evidence, rather than by the initial proposal score alone.

\section{Detailed Training Objectives}
\label{app:training_objective}

In the main paper, we summarize the training objective as:
\begin{equation}
    \mathcal{L}
    =
    \mathcal{L}_{\mathrm{out}}
    +
    \lambda_{\mathrm{sel}}\mathcal{L}_{\mathrm{sel}}
    +
    \lambda_{\mathrm{loop}}
    \left(
    \mathcal{L}_{\mathrm{mask}}
    +
    \mathcal{L}_{\mathrm{score}}
    \right).
\end{equation}
This section briefly describes the role of each term.

\paragraph{Language output loss.}
The language output loss \(\mathcal{L}_{\mathrm{out}}\) supervises the language-output pathway with the standard auto-regressive cross-entropy objective. It encourages the model to generate the response sequence containing valid task-related tokens, including \texttt{[SEG]} and \texttt{[LOC]}.

\paragraph{Candidate selection loss.}
The candidate selection loss \(\mathcal{L}_{\mathrm{sel}}\) supervises the Semantic-Spatial Selector to identify the target candidate from the candidate bank. We assign the candidate with the highest IoU to the ground-truth mask as the positive target:
\begin{equation}
    y_{\mathrm{cand}}
    =
    \arg\max_{i\in\{1,\dots,N_c\}}
    \mathrm{IoU}(c_i,m^{*}).
\end{equation}
The selector is trained with candidate-level cross-entropy supervision on the semantic branch, the spatial correction branch, and the final fused candidate score. We also include lightweight auxiliary constraints to improve text-candidate alignment and encourage complementary semantic and spatial representations.

\paragraph{Loopback verification loss.}
The loopback verification loss contains mask refinement supervision and mask-level score supervision. The mask loss \(\mathcal{L}_{\mathrm{mask}}\) supervises the refined point-level mask and the local region prior. Following common practice in segmentation, we use a combination of BCE, Dice, and Lovasz losses:
\begin{equation}
    \mathcal{L}_{\mathrm{mask}}
    =
    \mathcal{L}_{\mathrm{ref}}
    +
    \beta_{\mathrm{loc}}\mathcal{L}_{\mathrm{loc}},
\end{equation}
where both \(\mathcal{L}_{\mathrm{ref}}\) and \(\mathcal{L}_{\mathrm{loc}}\) are implemented with BCE, Dice, and Lovasz supervision. The refined mask loss \(\mathcal{L}_{\mathrm{ref}}\) is applied to the mask predicted from the target candidate, while the local region prior loss \(\mathcal{L}_{\mathrm{loc}}\) provides coarse local supervision around the target region.

\paragraph{Mask-level score supervision.}
The score loss \(\mathcal{L}_{\mathrm{score}}\) supervises the mask-level verification score in Loopback Verification. Among the selected Top-\(K\) candidates, the refined mask corresponding to the target candidate is treated as positive, while the remaining candidates are treated as negatives. This encourages the reranking module to select the refined mask with the best target consistency rather than relying only on the initial candidate score.
\section{Additional Ablation Studies}
\label{app:additional_ablation}

This section provides additional diagnostic ablation studies that are not included in the main paper.
These experiments include a diagnostic GT candidate bank analysis and an extended analysis of additional viewpoint information.

\subsection{GT Candidate Bank Analysis}
\label{app:gt_candidate_bank}

We further evaluate the model under a GT candidate bank setting to diagnose the effect of candidate coverage.
In this setting, the candidate bank is replaced by ground-truth object masks, while the model still needs to select the target candidate according to the language instruction.
Therefore, this setting removes errors from candidate generation but still evaluates the Semantic-Spatial Selector and Loopback Verification modules.
It is not used during standard inference and only serves as a diagnostic upper-bound analysis for candidate coverage.
The results are shown in Table~\ref{tab:gt_candidate_bank}.

\begin{table}[t]
\centering
\caption{GT candidate bank analysis on the ScanRefer validation set. In this diagnostic setting, the candidate bank is replaced by ground-truth object masks, but the model still predicts the final target by selecting from the candidate bank.}
\label{tab:gt_candidate_bank}
\small
\setlength{\tabcolsep}{8pt}
\begin{tabular}{lccc}
\toprule
Variant & Acc@0.25 & Acc@0.50 & mIoU \\
\midrule
Ours & 57.94 & 52.73 & 44.96 \\
GT candidate bank & 61.73 & 61.73 & 61.73 \\
\bottomrule
\end{tabular}
\end{table}

Since the candidates in this setting are ground-truth object masks, selecting the correct candidate usually yields an IoU close to 1, while selecting an incorrect candidate yields a very low IoU.
Therefore, Acc@0.25, Acc@0.50, and mIoU become nearly identical in this diagnostic setting.

\subsection{Viewpoint Variant}
\label{app:viewpoint_variant}

We further evaluate the effect of additional viewpoint information.
This setting provides the model with extra viewpoint cues and is therefore not used as a standard comparison in the main paper.
Instead, we report it in the appendix as an extended analysis.
The results are shown in Table~\ref{tab:viewpoint_variant}.

\begin{table}[t]
\centering
\caption{Effect of additional viewpoint information on the ScanRefer validation set.}
\label{tab:viewpoint_variant}
\small
\setlength{\tabcolsep}{8pt}
\begin{tabular}{lccc}
\toprule
Variant & Acc@0.25 & Acc@0.50 & mIoU \\
\midrule
Ours & 57.94 & 52.73 & 44.96 \\
Ours + viewpoint & 75.69 & 67.96 & 59.28 \\
\bottomrule
\end{tabular}
\end{table}

Since these viewpoint cues provide extra information beyond the standard input setting, this variant is not included in the main comparison.

\end{document}